\pdfoutput=1

\documentclass[11pt]{article}

\usepackage[preprint]{acl}

\usepackage{algorithm2e}
\usepackage{booktabs}

\usepackage{times}
\usepackage{latexsym}

\usepackage[T1]{fontenc}

\usepackage[utf8]{inputenc}

\usepackage{microtype}

\usepackage{inconsolata}

\usepackage{graphicx}
\usepackage{tikz}

\title{KréyoLID: From Language Identification Towards Language Mining}

\author{Rasul Dent$^1$, Pedro Ortiz Suarez$^2$, Thibault Clérice$^1$,  Benoît Sagot$^1$ \\ $^1$Inria, Paris, \texttt{\{firstname.lastname\}@inria.fr} \\ $^2$Common Crawl Foundation, Paris, \texttt{pedro@commoncrawl.org}}

\begin{document}
\maketitle
\begin{abstract}
Automatic language identification is frequently framed as a multi-class classification problem. However, when creating digital corpora for less commonly written languages, it may be more appropriate to consider it a data mining problem. For these varieties, one knows ahead of time that the vast majority of documents are of little interest. By minimizing resources spent on classifying such documents, we can create corpora much faster and with better coverage than using established pipelines. To demonstrate the effectiveness of the language mining perspective, we introduce a new pipeline and corpora for several French-based Creoles. 
\end{abstract}

\section{Introduction}

As Natural Language Processing (NLP) technologies gain prominence, so does the demand for corpora. Filtered versions of Common Crawl data, such as OSCAR \citep{suarezAsynchronousPipelineProcessing2019, abadjiUngoliantOptimizedPipeline2021, abadjiCleanerDocumentOrientedMultilingual2022}, MADLAD-400 \citep{kuduguntaMadlad400MultilingualDocumentlevel2024},  GlotCC \citep{kargaranGlotCCOpenBroadCoverage2024}, and Fineweb-2 \citep{penedo2024fineweb-2}, are one solution to this demand.  For English and Mandarin, this approach has yielded terabyte-sized datasets. 
Yet for most languages, even those written by millions of people, such as Romanized Arabic and Hindi, filtering Common Crawl has yielded  modest results.

As such, language identification (LID) for the thousands of varieities with little to no representation in web corpora has become a prominent research agenda \citep{caswellLanguageIDWild2020}. For example, \citet{kreutzerQualityGlanceAudit2022} document some types of noise that affect less common languages. Similarly, OpenLID \citet{burchellOpenDatasetModel2023}, GlotLID
\citet{kargaranGlotLIDLanguageIdentification2023}, and \citet{adebaraAfroLIDNeuralLanguage2022}, try to improve the coverage of LID models.



At the same time, Creole languages, which arose from intense language contact,  have garnered considerable attention \citep{lentWhatCreoleWants2022}. 
They generally have strong lexical overlap with a handful of widely-spoken languages, such as English, French, and Portuguese, but differ from them in morphosyntax. Although often lumped into a broad ``low-resource'' category, contact varieties bring unique challenges and opportunities for NLP \citep{birdLocalLanguagesThird2022}. 

For French-based Creoles (FCs), a group of 10-20 closely-related languages\footnote{The number depends on how we enumerate dialect chains.} spoken by approximately 15 million people, filtering web data remains challenging. These languages share many traits, but differ in sociolinguistic contexts, which has pronounced impacts on the effectiveness of NLP and especially LID. At one end of the spectrum lies Haitian Creole, which accounts for roughly 2/3 of FC speakers. It is the national language of Haiti and spoken by sizable diaspora communities in countries like the United States. As such, it has a robust online presence, and is well-represented in LID benchmarks like FLORES-200 \citep{nllb_teamNoLanguageLeft2022}. At the other end, the Creoles of Louisiana and Trinidad are critically endangered and have limited online presence. In the middle,  Lesser Antillean, French Guianese Creoles, Mauritian and Réunionese Creoles are partly institutionalized.\footnote{The French-based Creoles of the Lesser Antilles form a chain of closely-related and mutually intelligible varieties. Some of the main varieties in the chain are Guadeloupean, Dominincan,  Martinican and St. Lucian. Similar chains exist in the Republic of Mauritius (Mauritian proper, Rodriguan) and on Réunion Island (Créole des Hauts, Créole des Bas)} 

Although models  like  GlotLID
 and AfroLID \citep{adebaraAfroLIDNeuralLanguage2022}, try to take more FCs into account, even recent filters of Common Crawl snapshots, like GlotCC \citep{kargaranGlotCCOpenBroadCoverage2024} identify as few as 49 pages of content in Réunion Creole and 100 in Lesser Antillean Creole, each of which have vibrant online speech communities. 
 While such efforts have improved the situation for some languages, they still generally approach corpora creation 
 within a framework intended for the most common varieties, overlooking text distributions.



 More specifically, pipelines like Ungoliant \citep{abadjiUngoliantOptimizedPipeline2021, abadjiCleanerDocumentOrientedMultilingual2022} and GlotCC  break every document into sentence-to-paragraph-length segments. They then use multiclass classifiers, and especially the fastText architecture \citep{joulinBagTricksEfficient2017} on the resulting segments.  Yet, when we target a very small fraction of the Web, most data is of at best secondary importance and can be discarded with much less effort. 
 



With this in mind, we reframe (Creole) LID as a Needles-in-a-Haystack problem, and propose `Language Mining' as a solution. Our main claim is that we can efficiently identify a small French Creole cluster in large webcrawls by using a document-level Bag-of-Types strategy. To demonstrate this, we first introduce our threshold-based filtration system and then benchmark speed, recall, and false positive rate on clean Wikipedia data. Next, we estimate the recall capabilities on noisy web data by applying it to Creole subcorpora from three recent projects. After that,  we eliminate  99\% or more of distracting documents in a 2.6 billion page Common Crawl snapshot in a few hours on a medium-sized cluster while maintaining competitive recall. Finally, we explore additional passes and the remaining ``last kilometer problem'' of fine-grained LID. We will release filtered versions of Fineweb-2 and the results of first pass filtering on the December 2024 Common Crawl snapshot for each target label. Additionally, we will offer second pass corpora for Lesser Antillean and Mauritian Creoles which undergo further cleaning. \footnote{Pipeline available at \url{https://github.com/DEFI-COLaF/LanguageMining}. Corpora coming soon.}






\section{Related Work}
\label{Related Work}
We now review some of the most influential approaches to LID. Due to its important role in multilingual NLP, this task has a long history, which \citet{jauhiainenAutomaticLanguageIdentification2024} resume in depth. By the 90s, approaches based on n-gram frequencies,  including \citet{cavnarNgrambasedTextCategorization1994} had achieved 99\% accuracy on monolingual documents of sufficient length in several common languages.


However, subproblems like closely-related varieties, short texts, and code-switching showed that easy cases were only a fraction of the overall use cases for automatic language identification \citep{dasilvaIdentificationDocumentLanguage2006}. During  the late 2000s and early-mid 2010s, alternative approaches were explored, culminating in the adoption of sentence-level linear classifiers as a \textit{de facto} standard. In this section, we briefly review the usage of linear classifiers for language identification, and then explore how keyword methods provide a useful alternative.

\subsection{Linear Classifiers}

Since the late 2010s, the fastText algorithm \citep{joulinBagTricksEfficient2017} has been the basis of notable web-corpora pipelines. It combines a simple learned vector representation of words with a single hidden layer as well as a multiclass output layer, comparable to the Continuous Bag-of-Words model proposed by \citet{mikolovEfficientEstimationWord2013}. For efficiency, the output layer uses hierarchical softmax, which behaves similar to softmax but is much faster when there are many classes \citep{goodmanClassesFastMaximum2001}. \footnote{A one-vs-all option exists, but is less favored in LID.}



Besides fastText, Google's CLD3 utilizes many of the same principles, such as character n-grams, the hashing trick, and a shallow network consisting of an embedding layer, a single hidden layer, and a multiclass output layer \footnote{\url{https://github.com/ropensci/cld3} , see \citet{bothaNaturalLanguageProcessing2017}.}. However, CLD3 incorporates n-gram frequencies directly into the embedding layer, and the output layer uses softmax rather than hierarchical softmax. Given the similarities, CLD3 shares many of fastText's strengths and weaknesses, and is often used in baseline comparisons \citep{burchellOpenDatasetModel2023,kargaranGlotLIDLanguageIdentification2023}. 


\subsection{LID at Scale}
\label{LID}

\citet{graveLearningWordVectors2018} also introduce a corpus creation pipeline based on applying the fastText model to each line in a raw data dump, and then appending each line that passed a certain LID confidence threshold to the relevant corpus. Shortly thereafter, the OSCAR project refined this idea by parallelizing the transfer of data, and moving cleaning to occur before language identification \citep{suarezAsynchronousPipelineProcessing2019}, and then incorporating metadata and new optimizations \citep{abadjiUngoliantOptimizedPipeline2021}, ultimately resulting in the document-level Ungoliant pipeline \citep{abadjiCleanerDocumentOrientedMultilingual2022}.

Although OSCAR and older initiatives like C4 \citep{habernalC4CorpusMultilingualWebsize2016}  worked for the most common languages, they overlooked most other languages. \citet{caswellLanguageIDWild2020} detail conditions that cause models with high coverage under test conditions to be effectively unusable for many ostensible targets when applied at scale. Beyond obvious issues such as non-Unicode encoding, high-resource/out-of model related languages,  and short texts, there are mis-rendered PDFs, scripts mixed for visual effect, text with spaces between every character, and improbable repetitions of n-grams in high resource languages. For these issues,  they suggest post-filtering using lists words uncommon in the high-resource cousin, which can be arbitrarily precise. In addition, they also suggest self-supervised Transformer-based models, but these have the disadvantage of much slower runtimes. \citet{kreutzerQualityGlanceAudit2022} expand upon this work by identifying additional sources of error, such as inconsistent and incorrect use of language codes. 



\subsection{Creole LID}
As noted in the introduction, Creoles share numerous lexical affinities with other, generally more well-resourced, languages. 
This high lexical overlap, especially when combined with orthographic instability, makes LID particularly challenging  \citep{lentCreoleValMultilingualMultitask2024}. For example, \citet{caswellLanguageIDWild2020} show that even when ensembling complementary LID strategies, Naija (Nigerian Pidgin), remains one of the hardest languages to detect at scale due to its high overlap with English.

With the exception of Haitian Creole, LID for French-based Creoles has largely been pursued within the framework of models meant to detect at least one hundred languages. One relatively early work is \citet{sotoLanguageIdentificationGuadeloupean2020}, who focused on distinguishing Guadeloupean Creole among 103 languages using a fastText model. Somewhat later \citet{adebaraAfroLIDNeuralLanguage2022} took Mauritian and Seychellois into account using a neural model to distinguish between 517 African varieties, notably do not try to distinguish them from French or Haitian Creole. After, MADLAD-400 \citep{kuduguntaMadlad400MultilingualDocumentlevel2024}, built from multiple snapshots of Common Crawl, included several Creoles such as the French-based St. Lucian, Mauritian, and Seselwa, and the English-based Eastern Maroon and Belizean Creoles among their 419 languages. At the time of writing, GlotLID \citet{kargaranGlotLIDLanguageIdentification2023} offers the most extensive coverage across Creole languages, and has recently been used to create two large scale corpora, GlotCC \citep{kargaranGlotCCOpenBroadCoverage2024} and FineWeb-2 \citep{penedo2024fineweb-2}, which will serve as points of reference for our system and are further described in Section \ref{Recall on Large Datasets}.

 \subsection{Keyword Search}
 \label{Keyword Search}
 
 Keyword methods date back to the early days of LID \citep{pragerLinguiniLanguageIdentification1999, jauhiainenAutomaticLanguageIdentification2024}.  An important example was the Crúbadán Project, which revolved around modeling languages as search queries for a custom web crawler \citep{scannellCrubadanProjectCorpus2007}. 
 More specifically, they identified languages like Irish by searching combinations of at least one common, yet distinctive stopword and at least one other word common to the language using the `AND' and 'OR' operations. When reliable wordlists were on hand, they could be used directly. In other cases, lists could be bootstrapped from trusted corpora (or fluent speakers). For validating the results of the query and outgoing links, the crawler augmented simple character trigram frequencies with basic metadata about the relevant languages, such as languages they are likely to co-appear with.\footnote{For example, pages with content in Lingala would likely also contain French, and English might skew any language.} 

 More recently, \citet{lauExtractionFinegrainedClassification2024} used key-string methods to distinguish written Cantonese, Standard Written Chinese, and intermediate varieties (either mixed or unmarked). They emphasize that focusing on string operations allows for LID decisions to be made up to 4x faster than the commonly used fastText lid.176 model. Similarly, the Molyé project implemented a keyword-based approach to identify historical examples of nonstandard French-related varieties \citep{dentMolyeCorpusbasedApproach2024a}.  

\section{Language Mining}
\begin{figure}[htbp]
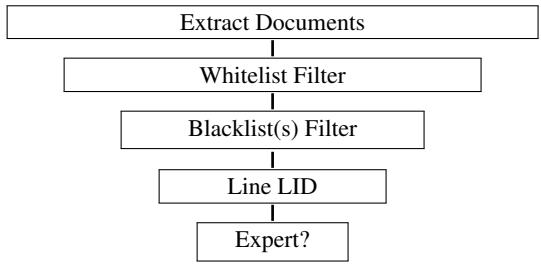

\begin{center}\small
~\\
~\\
\framebox[7cm]{Extract Documents}\\
\rule{1pt}{1.5ex} \\
\framebox[5.5cm]{Whitelist Filter}\\
\rule{1pt}{1.5ex} \\
\framebox[4cm]{Blacklist(s) Filter}\\
\rule{1pt}{1.5ex} \\
\framebox[3cm]{Line LID}\\
\rule{1pt}{1.5ex} \\
\framebox[2cm]{Expert?}\\

\end{center} 
\caption{Language Mining Pipeline} \label{Language Mining schema}
\end{figure}

The phrase ``needle in a haystack''  is a longstanding metaphor for search scenarios where one or more target ``needle(s)'' is/are embedded in a very large number of distracting ``hay'' records \citep{crammerNeedleHaystackLocal2004}. 
 The heart of a Needle(s)-in-a-Haystack problem is to eliminate hay quickly. For language identification, this means ignoring documents that lack (quasi)-unique features of our target language(s). To find the few documents that \textbf{do} contain clusters of such features, we create a kind of search engine that uses lexicons as queries and returns ranked document- and line-level corpora. Conceptually, this is similar to the Crúbadan Project. Yet, whereas search engines combine several distinct components, including a crawler, an indexer, a scoring/ranking algorithm, and a user interface, our focus is limited to indexing and scoring data crawled by a third-party.\footnote{In this case, the Common Crawl Foundation.} To implement this data triage, we introduce an indexing-scoring system that operates in two phases: first at the document-level, and then at the sentence level. The main components of this method, which we call `Language Mining', are shown in Figure \ref{Language Mining schema} and expanded upon in this section.


\subsection{Document-level}
Since we want to identify a very small portion of the overall data, we take an approach of staggered filtration, where any example that makes it through to the final corpus will have to make it through all of the filters. Importantly, filters can have very different runtimes. The task then is to  order them according to mean runtime per example rejected, while repeating as little work as possible. Fortunately, lexicon-based filters are a very fast and (potentially) high-precision way to establish that a document likely contains data in a target language. 


\subsubsection{Feature Extraction (Tokenization)}
\label{Tokenization}
Wordlist methods do not inherently require tokenization. Through hashing, they can be generalized to work with approach to work with any substring. For the prototype, however, we focus on whitespace tokens, which align closely with words in the case of FCs (and many, but not all, other languages), because this allows us to use fast built-in string methods. For languages with distinct orthographies and relatively little bound  morphology, as is the case for several French-based Creoles, the wordlists provided by \citet{caswellLanguageIDWild2020} are a very convenient point of entry. In Section \ref{Speed Results}, we briefly consider the impact punctualization normalization as well. For languages that are heavily inflected or written in a script that relies less on whitespace, fixed-length character windows are a fast, generalizable, and readily available alternative \citep{google-2025-fun}.


\subsubsection{Whitelists and Blacklists}
Methods depending on a few key features tend to capture noise characterized by the same features. Line level lexical approaches are particularly vulnerable to collisions  \citep{caswellLanguageIDWild2020, kreutzerQualityGlanceAudit2022}.  However, this problem is much easier to manage at the document level, because we can also make blacklists that characterize different kinds of noise to use as additional filters. While this strategy will not eliminate all of the noise, it is effective for eliminating various kinds of spam, especially adult websites. In some scenarios, distractor languages can also be filtered with blacklists \citep{ljubesicLanguageIndentificationHow2007}.

Additionally, after the first pass, we can inspect the ranked data to identify specific sources of reliable or confusing data. For closely-related languages, this is an easy to way capture human intuitions that may be difficult to model succinctly.


\subsubsection{Scoring, Ranking and Indexing}
\label{sec:sco-ran-ind}

To identify candidate documents, we first calculate both a whitelist score (wsc). This score can be computed in two ways: using a boolean match per type, or using type or token frequencies. Additionally, one could follow \citet{scannellCrubadanProjectCorpus2007} and make certain types mandatory. For the initial implementation, we use simple the sum of boolean matches, as this does not require calibrating relative frequency information. For efficiency, we store each list as a set and calculate this score as the intersection of the list and the document's token-types.

Having calculated whitelist scores, we exclude the documents whose scores are below a certain value (threshold). For the documents that pass this threshold, we then calculate a blacklist score (bsc) using the same scoring mechanism, and eliminate scores above a second value called the tolerance. Next, we rank the remaining documents by whitelist score. This allows us to prioritize the highest scoring ones, which are likely either in the target language or special edge cases (see Algorithm \ref{algo:one}).



If we have sufficient space and are interested in eventually exploring other languages, we can greatly reduce the runtime of future searches by saving the vocabularies of all (or most) documents as indices. When space is limited and/or we are sure that we are only interested in specific languages, we only need to save the highest scoring documents.

\begin{algorithm}[htp!]\small
\captionsetup{labelfont={sc,bf}, labelsep=newline}
\caption{Double List Filtering Algorithm}
\label{algo:one}
\SetAlgoLined
\SetKwData{Documents}{D}\SetKwData{Wordset}{W}\SetKwData{Blacklist}{B}\SetKwData{Threshold}{threshold}\SetKwData{Tolerance}{tolerance}\SetKwData{Tokens}{tokens}
\SetKwData{Types}{types}\SetKwData{WSC}{wsc}\SetKwData{BSC}{bsc}

\SetKwFunction{Tokenize}{tokenize}\SetKwFunction{Map}{map}
\SetKwFunction{Len}{len}\SetKwFunction{intersection}{Intersection}\SetKwFunction{Score}{score}\SetKwFunction{Save}{save}
\SetKwFunction{Set}{set}

\SetKwInOut{Input}{input}\SetKwInOut{Output}{output}
\Input{W whitelist, B blacklist(s), D documents}


\BlankLine
\For{$i \leftarrow 0$ \KwTo $\Len{\Documents}$}{

\emph{First get token-types}

$\Tokens \leftarrow \Tokenize {\Documents}$\;
$\Types \leftarrow \Set{\Tokens}$\;

\BlankLine
\emph{Then score}

$\WSC \leftarrow \Score{\Types, \Wordset}$\;

$\emph{Optionally cache all vocabularies}$

\BlankLine
\emph{Then filter}

    \If{$\WSC[i]$ $\ge$ \Threshold }{
    $\BSC \leftarrow \Score{\Types, \Blacklist}$\;
        \If{\BSC $<$ \Tolerance }
        {\Save {$(\Documents[i], \WSC[i])$}

        }
    }

}
\BlankLine
\emph{Sort saved documents by score (descending)}
\end{algorithm}

    


\subsection{Line-level}
\label{Line-level filtering}
At the line-level, we have a wide range of possible filters. The simplest of all is a length check, which can help remove common boilerplate \citep{kohlschutterBoilerplateDetectionUsing2010}. However, this may exclude list-based content like dictionaries. Script checks and line-level keyword filters are slightly more expensive, but still lightweight options. Beyond this, we can still classify sentences using more intensive models, such as fastText or even Transformers. The main difference between the staggered filtration approach and the more established pipelines is that we aim to only use the intensive models when needed. 

In this work, we explore the extent to which line-level filtering can also be performed with wordlist-based approaches. For ranking lines, we start with a type-based score, like at the document-level. However, observing that naive line-tokenization on less-structured data can yield extremely long lines, we introduce normalization normalize the score by dividing by the length of the string. This allows short but meaningful strings with types to appear at the top of the pile while pushing long strings that contain a few target types by chance to the bottom. Duplicates also cluster together when ranked, and can be removed if desired.



\section{Measuring Performance}
Given the sociolinguistic situations mentioned in the introduction, we focus on improving LID for Lesser Antillean Creole(s) \textit{acf} and \textit{gcf},\footnote{The ISO codes \textit{acf} and \textit{gcf} are officially listed as St.~Lucian and Guadeloupean, respectively, Within the Lesser Antilles, Guadeloupe has several distinct features, hence \textit{gcf}, but the grouping of other islands,  under these two labels varies.} French Guianese \textit{gcr}, and Mauritian \textit{mfe}, Seychellois \textit{crs}, and Réunionese \textit{rcf}.  While do not include endangered varieties because the reference corpora do not address them. As Haitian Creole is already covered by many systems, we include it mainly as a point of reference.

Comparing our method with other systems is difficult for two reasons. Firstly, building LID test sets for long-tail languages is complicated in general, since the data available are often overly clean, skewed to a few reliable sources, and/or themselves collected with LID. Secondly, because our primary focus is document-level filtration, sentence-level datasets like FLORES-200 are likely to severely underestimate our performance. 

With this in mind, we adopt a four-pronged approach. First, we first create a benchmark to compare the speed, recall, and false positivity rate of Language Mining and GlotLID. To get an idea of how much content we should expect to find in a snapshot, we then apply our document-level filtration system to subcorpora from three recent, large-scale filtered corpora that feature our target languages, namely MADLAD-400, GlotCC, and Fineweb-2. For all three corpora, we report our recall on the `clean' portions of these corpora, as well as our ability to find usable data in the discard portion of the largest, Fineweb-2. After that, we test document-level filtering on a full Common Crawl snapshot with different wordlists at two thresholds.  Lastly, we briefly explore the potential of type-based filtering for the second pass, and qualitatively estimate corpora quality.

\label{Intrinsic Evaluation}

\subsection{Wikipedia-based Benchmarking}
\label{Benchmarking}
For benchmarking the speed and language mining performance of our approach, we create a dataset of 10,000 Wikipedia entries, with 9800 articles in French and 200 in French Guianese Creole, taken from their respective Hugging Face Wikipedia datasets.\footnote{\url{https://huggingface.co/datasets/wikimedia/wikipedia}.} To reduce the chance of Creole data appearing in the French portion, which could potentially lead to true positives being mistaken for false positives, we skip articles that contain the word ``créole''. 
We compare our Language Mining approach to the full GlotLID model \citep{kargaranGlotLIDLanguageIdentification2023}, which covers over 1600 linguistic varieties, as well as a version of GlotLID with the output space reduced to 3 languages (French, French Guianese Creole, English). We compare these two GlotLID models' speed and performance to those of Language Mining when used to mine data for a single language (`Min-1' setting) as well as for 3 languages (French Guianese, Lesser Antillean, Mauritian) simultaneously (`Min-3' setting), based on our wordlists for the respective languages. To convert line-level fastText predictions into document-level labels, we implement the Ungoliant document-scoring procedure \citep{abadjiCleanerDocumentOrientedMultilingual2022} adopted by \citet{kargaranGlotCCOpenBroadCoverage2024}. We also explore both the impact of regex-based punctuation normalization, as mentioned in Section \ref{Tokenization}, and the impact of the value of the score threshold defined in Section~\ref{sec:sco-ran-ind}.

\subsection{Recall on Large Datasets}
\label{Recall on Large Datasets}


MADLAD-400 \citep{kuduguntaMadlad400MultilingualDocumentlevel2024} is the oldest of the three corpora, with a cutoff date of August 1st, 2022. Built from all Common Crawl snapshots up to that point, it has sizeable subcorpora forfive of our target Creoles: Lesser Antillean, Haitian, Mauritian, Réunionese, and Seychellois. Since the Transformer-based LID system used to create MADLAD-400 is not open-source, we are unable to directly compare its speed, recall or false positivity rate to those of our model.

GlotCC \citep{kargaranGlotCCOpenBroadCoverage2024} combines the Ungoliant pipeline with  GlotLID, and covers the February/March 2024 snapshot and portions of the September/October and November/December 2023 ones. In addition to the five target languages covered by MADLAD-400 GlotLID and GlotCC also cover Guadeloupean and French Guianese Creoles. Although GlotCC is smaller than the other two, it also comes with details about speed..

Fineweb-2 \citep{penedo2024fineweb-2} uses GlotLID to cover our targets. Like MADLAD-400, it is built from all Common Crawl snapshots released prior to its cutoff date of April 2024 (96 in total). By using a high coverage model on several petabytes of raw data, Fineweb-2 has created appear what, to our knowledge, are the largest \textit{open} web-scraped corpora for several FCs, and thus represents the state-of-the-art at the time of writing.

\subsection{Filtering Common Crawl}
We use the December 2024 (CC-MAIN-2024-51) Common Crawl snapshot, which amounts to 21 TB of raw data, to test Language Mining at scale. For development purposes (mainly exploring thresholds and tolerances as defined in section \ref{sec:sco-ran-ind}), we also used subsets of the July 2024 snapshot. 
During prototype, we found that parallel document-level sorting in Rust was nearly twice as fast as Python. For an efficiently parallelized first pass on the full 21 TB, we used the Rust implementation and further configured the program for a SLURM system.  In \ref{Speed Results}, we report the runtime on two kinds of node, Broadwell Xeon e5-2650 v4 processors and Broadwell Xeon e5-2695 v3/v4. Here, we include the time needed to parse WET files into documents, but, as with GlotCC and OSCAR  do not consider data transfer and data compression, which are beyond the scope of the LID algorithm. In addition to document-level filtering, we also rank each line as described in Section \ref{Line-level filtering}. This isolates quotes embedded in other languages and penalizes long strings of noise. To explore subsequent passes on smaller data, we returned to prototyping in Python. 


\subsection{Qualitative Evaluation}
 Due to the complications mentioned in the beginning of this section, we focus on qualitative estimations of corpora quality and diversity. After the first pass, one of the authors familiar with the relevant languages manually examined the resulting document and line level corpora, along similar lines to the audits of \citet{kreutzerQualityGlanceAudit2022}. This provided insights which were taken into account for the second pass on two target languages: Lesser Antillean and Mauritian. These two then received a second round of qualitative evaluation.

\section{Results}
\subsection{Wikipedia-based Benchmark Results}
\label{sec:wikiresults}
As mentioned in \ref{Benchmarking}, we first compare the speed of GlotLID and Language Mining.\footnote{We run these experiments in serial with an Intel i7-1370P.} Language Mining takes 0.46s for a single target language (Min-1)  and achieves 79.0\% recall with a base threshold of 5 disjunctive types per document (Table \ref{fig:speed_benchmark}). Adding two languages (Min-3) only adds 0.05s on average compared to Min-1, and has no effect on recall or false positivity rate (FPR) because the scores for each language are independent. Punctuation-aware tokenization (Min-1-P) nearly doubles the runtime for a single language and increases recall by 0.5\% (a single document) but doubles the FPR. Using the full GlotLID model (Glot-Full) takes over 200 times longer than Min-1, but achieves a high 91\% recall. Since GlotLID was trained in part on Wikipedia, this high performance is to be expected. After restricting the output space of GlotLID to 3 languages (Glot-3), speed improves, but remains 45 times slower. Recall rises to 100 \% and FPR falls to 0. 


\begin{table}[h]
    \centering\small

    \resizebox{\columnwidth}{!}
    {
    \begin{tabular}{lrrrrr}\toprule
     Model & Glot-Full & Glot-3 & Min-1 &Min-3 & Min-1-P\\
    \midrule
     Mean Speed (s) & 114.27 & 21.45 & 0.46 & 0.51 & 0.83\\
     \textit{gcr} Recall \% &  91.00  & 100.00     &   79.00 &   79.00  &  79.50  \\
     \textit{gcr} FPR \% & 0.0   & 0.0     & 0.04  &  0.04 &  0.09    \\
     \bottomrule
\end{tabular}
    }
    \caption{Benchmark on the mixed Wikipedia corpus.}
    \label{fig:speed_benchmark}
\end{table}

When we test Min-1 at different score thresholds (Table \ref{fig:recall_benchmark}), we see a gradual decrease in recall as the required number of disjunctive types approaches 10. With a conservative threshold of 5 types, we are able to eliminate almost all of the French documents, while keeping nearly 80\% of the Creole articles. Beyond a threshold of 10, performance continues to decrease, but without additional benefits on the two-language dataset. Overall, we validate the wordlist approach of \citet{caswellLanguageIDWild2020} for quickly eliminating high-resource cousins using choosable thresholds.


\begin{table}[h]
    \centering\small
    \begin{tabular}{lrrrrrr}\toprule
     Threshold   &  1 & 3 &  5 & 10 & 15 &\\
    \midrule
     True Positives  & 197.0 & 176.0 & 158.0 & 118.0 & 44.0 \\
     False Positives  & 1068.0 & 38.0 & 4.0 & 0.0& 0.0\\
     Recall \%  & 98.5 & 88.0 & 79.00 & 59.0 & 22.0 \\
     
     FPR \%  &  11.6 & 0.5 & 0.1 &0.0  & 0.0\\

     \bottomrule
\end{tabular}
    \caption{The effect of threshold on recall and false positive rate (FPR) for French Guianese Creole}
    \label{fig:recall_benchmark}
\end{table}
    


\subsection{Recall on Other Corpora}
 In Table \ref{fig:recall_on_clean}, we show that, with a threshold value of 5,  recall is well over 95\%  for most of the `clean' datasets, which shows that our method is quickly finds documents that would score well on slower models. For GlotLID-acf, the corpus size was very small (6). We verified that the last document was noise, and thus our algorithm correctly excluded it. This issue was most pronounced with the FineWeb-2-mfe clean subcorpus, where most of the raw data appears to be noise of the ``repetitive ngram'' type, especially from commercial websites.

\begin{table}[h]
    \centering\small
    \begin{tabular}{lrrr@{~~~~~~}c@{~~~~~~}rrr}\toprule
     Corpus & acf & gcf &gcr & hat & crs & mfe & rcf \\
    \midrule
     GlC &   83.3& 100.0 & 100.0 & 100.0 & 100.0 & 95.5 & 95.9 \\
     MAD &  100.0 &NA & NA & 99.9& 99.5 & 98.1 & 99.0 \\
     FW2 &   77.9 & 99.8 & 98.5 & 99.3 & 89.1& 38.0 &93.7  \\\bottomrule
\end{tabular}
    \caption{Recall percentages on comparable corpora}
    \label{fig:recall_on_clean}
\end{table}

We also applied our type-score filtering to
the `removed' subcorpora of Fineweb-2, which consist of data that received a target language label, but was of questionable quality. As shown in 
Table~\ref{tab:fineweb2}, the `removed' piles range from roughly the same size as the `clean' piles in the case for Réunionese Creole (rcf) to 66x bigger for Seychellois Creole (crs). When we filter the `removed' documents, we find that even though passing documents are but a small percentage of the `removed' data in several cases, they nearly double the Mauritian (mfe) and Haitian (hat) corpora, and are over 9 times more numerous than the filtered `clean' documents for Lesser Antillean Creole (acf). Thus, beyond post-filtering, we can pre-filter noise too.

\begin{table}
    \centering\small
     \resizebox{\columnwidth}{!}{
    \begin{tabular}{lrrr@{~~~~~~}r@{~~~~~~}rrr}\toprule
     Language & acf & gcf &gcr & hat & crs & mfe & rcf \\
    \midrule
     Clean-raw & 1.1 & 2.8  &0.9 & 224.4 &3.5  &20.4 &7.9  \\
     Rem-raw  & 109.0 & 10.9 &5.6  & 4466.7 &233.9 &807.0  &7.8 \\
     Clean-filt & 0.9 &2.8 &0.9 & 222.8 &3.1  & 7.8 &7.4  \\
      Rem-filt  & 7.4 & 3.0 &1.8  &  351.0 &2.5 &6.9  &3.5 \\
      Total-filt  & 8.3 & 5.8 &2.7  &  573.8 &5.6 &14.6  &10.9 \\\bottomrule
\end{tabular}
}
    \caption{Documents (thousands) in Fineweb-2 Corpora clean and noisy subcorpora before and after filtering}
    \label{tab:fineweb2}
\end{table}


\subsection{Common Crawl}
\subsubsection{First Pass}
\label{Speed Results}
Using the configuration described in Section~\ref{Intrinsic Evaluation}, we ran the pipeline several times to measure the impact of different wordlists and thresholds (5 and 10). Overall, we were able to index  21 TB of raw data using 9 parallel jobs with 32 CPUs in 2 to 4.5 hours of wall time, with the exact runtime depending on the cluster. More specifically, on nodes with Broadwell Xeon e5-2650 v4 processors consistently finished their respective jobs in two hours, which translates to an average speed of roughly 1258 pages per CPU per second.\footnote{$2.6 \times 10^9$ webpages $/\ (9$ jobs $\times\ 32$ cpu$/$job $\times\ 2$ hr $\times\ 3600$ s$/$hr$)$.} On nodes using other processors, including Broadwell Xeon e5-2695 v3/v4, were closer to the 4-hour mark, but it is also possible that external factors within the cluster affected the exact runtime. However, even when jobs were not simultaneous due to cluster conditions, the pre-filtering stage ran at least 50-100x faster than GlotCC's reported wall time of 340~hours \citep{kargaranGlotCCOpenBroadCoverage2024}. This is congruent with our results in Section~\ref{Benchmarking} especially if we note that a considerable percentage of our actual runtime is merely reading the input data. 

The choice of wordlist and threshold did not seem to impact runtime. However, there were appreciable differences in the sizes of the indexed corpora. As seen in Table \ref{fig:indexed_size}, the Lesser Antillean (acf, gcf) and French Guianese (gcr) lists yield corpora 2-3x smaller than the Haitian and Indian Ocean lists, which is likely because the former use more  diacritics.

\begin{table}[h]
    \centering\small
    \begin{tabular}{lrrr@{~~~~~~}r@{~~~~~~}rrr}\toprule
     Language & acf & gcf & gcr & hat &  crs & mfe & rcf \\
    \midrule
     5 & 158 & 93 & 116  & 265 & 212 & 260 & 233 \\
     10 & 13 & 12  & 15 & 34 & 31 & 39 & 46\\\bottomrule
\end{tabular}
    \caption{Indexed size (GB) by language and threshold}
    \label{fig:indexed_size}
\end{table}

\subsubsection{Second Pass}
 Once the documents and lines are ranked, the highest-scoring content (for document, from the mid teens upward to hundreds) is reliably in either the target language or a closely-related sister language. Medium scores are sometimes true positives, but a substantial amount of content is in a handful superficially similar distractor languages, such as Catalan, Roman Hindi, and English with erratic spacing. At the document level, low scores are often indicative of noise, but sometimes come from short quotes, either as standalone documents, or embedded in a longer work in another language. At the line level, however, low normalized scores are particularly indicative of noise. 

Due to the sort operation, the second pass is run as a single job, which makes identifying a (higher and language specific) data loading threshold important for speed. Focusing on one language from the Americas (Lesser Antillean) and one language from the Indian Ocean (Mauritian), the second pass on the former takes less than an hour with a loading threshold of 10, while the latter needs a threshold of 14 for similar speed (compare with Table \ref{fig:indexed_size}). We easily remove less-closely related languages at the document-level by using the WARC-identified language. Languages removed this way include Swedish, Romanian, and Turkish. To remove data from closely-related FCs, namely French Guianese, Haitian, and Réunionese, we eliminate documents where they score higher than the target, and also filter by source. For example, the first pass for both corpora includes pages from French Guianese Wikipedia and a specific Réunionese newspaper,  which can be removed based on the URL. 

\subsection{Qualitative Evaluation}
 A small portion of the removed-filtered data from Fineweb-2 consists of false positives in languages such as Hilgaynon and phonemically-spelled French. However, the majority are in the correct language or at the very least, the correct subgroup (Americas or Indian Ocean). In broad terms, the content found within each language corresponded with what \citet{robinsonKreyolMTBuildingMT2024a} found through semi-manual collection. FBible translations were very prominent in the Lesser Antillean corpora, but many song lyrics where also found. French Guianese data was predictably mainly from Wikipedia. Large amounts of the Mauritian corpus came from a single prolific language activist, but religious texts, news, and music were also detectable. Réunionese data was similarly dominated by one local newspaper, but web forums and cultural content had a sizable presence. Seychellois data was particularly diverse; parliamentary reports were predictably well represented, but we also found full-length linguistic studies and the other genres already mentioned. Curiously, we found aligned Bible translations for unexpected languages, including Amharic, Arabic, Toba Batak, Biak, and Ghomálá' \footnote{Amharic, Toba Batak and Biak, and Ghomálá' are spoken in Ethiopia, Indonesia and Cameroon, respectively.} in the Indian Ocean datasets.
 
 For our Common Crawl filter, the range of content is similar to what is observed in Fineweb-2. Low scoring data is numerically dominant after the first pass, but simple visual inspection suffices to identify a cutoff point beyond which above which the signal-to-noise ratio increases dramatically. Ultimately, the exact subcorpora sizes depend on our threshold, but for both languages, we find several hundred reliable documents even at high (>30) thresholds. We will release our filtered versions of FineWeb-2, which significantly increase the amount of readily-usable web data for several languages, as well as the intermediate outputs of the first pass for all varieties and the aforementioned second pass datasets for further LID research.

\section{Conclusion}
In summary, we have introduced Language Mining as an efficient approach to web corpora creation for less commonly written languages. Taking advantage of the large class imbalance among varieties, we are able to eliminate the overwhelming majority of documents by counting how many distinctive types appear in the document. For French-based Creoles, such types can be identified quickly using whitespace tokenization. Rapid LID facilitates new kinds of exploration of crawled data, including but not limited to training language models.

\section{Limitations}

\subsection{Scaling Up}
As mentioned in Section \ref{Intrinsic Evaluation}, it is common to report  only the time spent on LID, ignoring the transfer and decompression of data. Thus far, we have take a similar approach. However, since our Language Mining approach runs several in such little time, one might wonder why we did not process multiple snapshots. Effectively, improving the efficiency of LID causes the entire pipeline for rarer varieties to no longer be bound by CPU-intensive inferences, but rather internet bandwidth.

More specifically, it took 19.25 hours to copy one snapshot (7.37TB compressed) from AWS storage (USA) to our compute cluster (Europe) using the Common Crawl Downloader with the suggested 10 parallel threads \footnote{https://github.com/commoncrawl/cc-downloader}. Although we were able to distribute the decompression across several nodes, it took an additional 8 hours of wall time to decompress the dataset. Yet, as we have shown in \ref{Speed Results}, we ideally only look at most of the data once (to throw it away), which can be accomplished in a few hours using resources of a similar scale. While copying the raw data is still useful for development purposes, such as debugging very rare conditions, it is likely necessary to process the data \textit{in situ} to truly take advantage of the speed of the method.

\subsection{Beyond French-based Creoles}
Although the naive Bag-of-Types approach is very effective for varieties that use distinct orthographies, as is this case with the more standardized of the French-based Creoles, there are typologically similar languages for which a whitespace-type based approach is less effective, and there are two obvious reasons for this.

Because the wordlists provided by \citet{caswellLanguageIDWild2020} were designed to be used as second-pass filters, some words are only or two characters long. In these cases, a wordlist-based method is particularly vulnerable to 'A N T S P E A K', where texts are broken up by excessive whitespaces. In this work, we mitigated this drawback by introducing a minimum character length when constructing our wordsets. 

However, certain languages , notably including Nigerian Pidgin/Naijá, appear to have few distinct, yet common types. In the provided `pcm' wordlist, many of the ostensibly disjunctive words are standard English words that deal with topics like crime and pop culture. While an obvious first step is to simply remove such words, Naijá and closely related varieties like Jamaican Patois show that we cannot complete dispense with syntax. With this in mind, we also an exploratory regex-based experiment to complement the main work. The preliminary analysis suggests that taking a small number of multiword expressions into account, such as \textit{no dey}, can improve coverage for these languages. Thus, a generalizable way to cross whitespace boundaries without wasting time on superfluous n-grams would be a welcome improvement.  

\section*{Acknowledgements}

This work was primarily funded by the Inria ``Défi''-type project COLaF. This work was also partly funded  by the last author's chair in the PRAIRIE institute funded by the French national agency ANR as part of the ``Investissements d'avenir'' programme under the reference ANR-19-P3IA-0001.

\bibliography{custom}




\end{document}